\documentclass[10pt,twocolumn,letterpaper]{article}

\usepackage{cvpr}
\usepackage{times}
\usepackage{epsfig}
\usepackage{graphicx}
\usepackage{amsmath}
\usepackage{amssymb}
\usepackage{mathtools}
\usepackage{textcomp}
\usepackage{titling}

\date{}

\usepackage[pagebackref=true,breaklinks=true,letterpaper=true,colorlinks,bookmarks=false]{hyperref}

\cvprfinalcopy 

\def\cvprPaperID{2446} 

\ifcvprfinal\fi
\begin{document}

\title{Improving Consistency and Correctness of Sequence Inpainting using Semantically Guided Generative Adversarial Network} 

\author{Avisek Lahiri\thanks{Denotes equal contribution.}~$^1$\hspace{5mm} Arnav Kumar Jain$^{*}$~$^2$ \hspace{5mm} Prabir Kumar Biswas$^3$ \hspace{5mm}  Pabitra Mitra$^4$\\
Indian Institute of Technology Kharagpur\\
{\tt\small \{$^1$avisek, $^3$pkb\}@ece.iitkgp.ernet.in, \{$^2$arnavkj95, $^4$pabitra\}@iitkgp.ac.in}
}

\maketitle

\begin{abstract}
Contemporary benchmark methods for image inpainting are based on deep generative models and specifically leverage adversarial loss for yielding realistic reconstructions. However, these models cannot be directly applied on image/video sequences because of an intrinsic drawback- the reconstructions might be independently realistic,  but, when visualized as a sequence, often lacks fidelity to the original uncorrupted sequence. The fundamental reason is that these methods try to find the best matching latent space representation near to natural image manifold without any explicit distance based loss. In this paper, we present a semantically conditioned Generative Adversarial Network (GAN) for sequence inpainting. The conditional information constrains the GAN to map a latent representation to a point in image manifold respecting the underlying pose and semantics of the scene. To the best of our knowledge, this is the first work which simultaneously addresses consistency and correctness of generative model based inpainting. We show that our generative model learns to disentangle pose and appearance information; this independence is exploited by our model to generate highly consistent reconstructions. The conditional information also aids the generator network in GAN to produce sharper images compared to the original GAN formulation. 
This helps in achieving more appealing inpainting performance. Though generic, our algorithm was targeted for inpainting on faces. When applied on CelebA and Youtube Faces datasets, the proposed method results in a significant improvement over the current benchmark, both in terms of quantitative evaluation (Peak Signal to Noise Ratio) and human visual scoring over diversified combinations of resolutions and deformations.
\end{abstract}

\begin{figure}[!t]
 \centering
 \includegraphics[scale = 0.8]{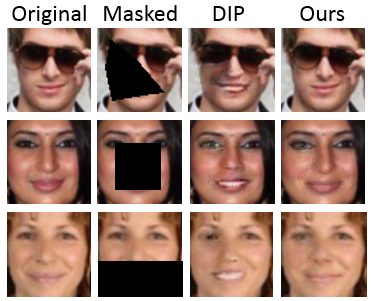}
 \caption{Exemplary success of our model in simultaneously preserving facial semantics(appearance and expressions) and improving inpaiting quality. Benchmark generative models such as DIP \cite{yeh2017semantic} are agnostic to holistic facial semantics and  thus generate independently realistic, yet structurally inconsistent solutions.}
 \label{fig_problem_actual_shown}
 \end{figure}
\begin{figure}[!t]
\centering
\includegraphics[scale = 0.35]{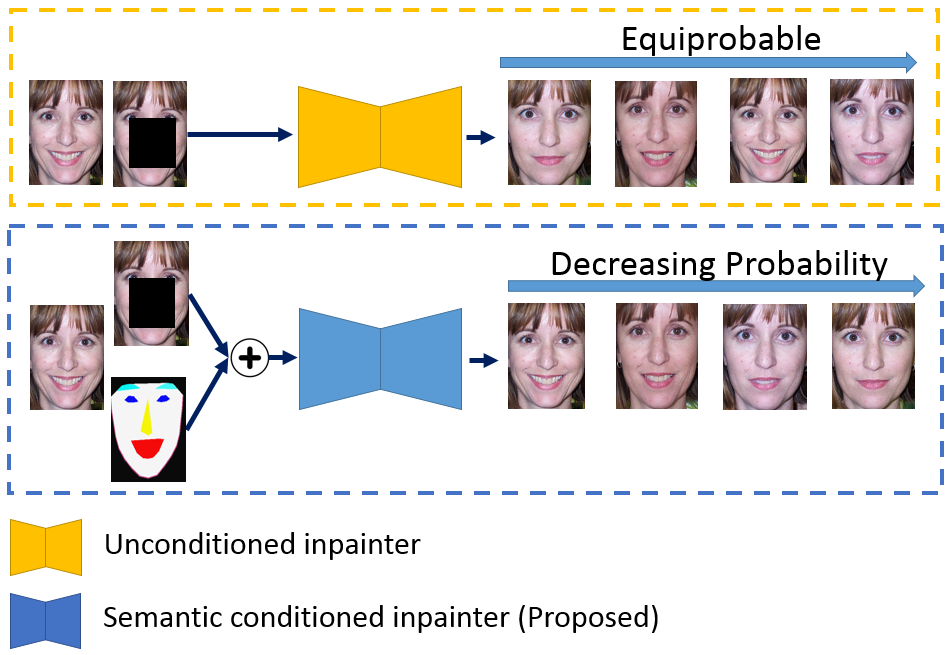}
\caption{Illustration of multi model image completion possibility of  GAN based inpainting methods. Given a corrupted image, an unconditioned inpainting algorithm such as \cite{yeh2017semantic} samples from a uniform distribution of viable inpainted images. However, if conditioned by facial semantics, the sampling distribution is biased towards samples which preserve original facial semantics.}
\label{fig_problem}
\end{figure}
%
\section{Introduction}
Semantic inpainting refers to reconstructions of damaged portions of an image using available neighborhood information. In this paper, we are interested in investigating the role of automated dense semantic conditioning to generative adversarial networks (GAN) \cite{goodfellow2014generative} for the specific task of semantic inpainting. We have focused on the special case of semantic inpainting of faces because faces are tough to inpaint due to presence of finer semantic details. Also, due to contemporary proliferation of multimedia services, video calling is deemed to become a frequent mode of communication and in such video streams, human faces occupy major part of a frame. Thus computer vision guided facial sequence inpainting is the call of the hour. 
Specifically, we wish to study and improve upon two aspects, viz., a) consistency and b) correctness.  Consistency is applicable in case of sequence inpainting, in which we measure the coherence among a group of reconstructed frames. If not accounted for, generative models render abrupt structural changes and unpleasing flickering  effects over stationary portions of frames. This is an intrinsic nature of generative model because the forward process of mapping a corrupted section to a valid image manifold is multimodal. An illustration is shown in Figure \ref{fig_problem} (Refer to Figure\ref{fig_problem_actual_shown} for actual comparison of outputs), wherein a generative model has multiple independent and equiprobable  options to semantically fill in the corrupted portion of the image. However, if the model is applied on a stream of video frames, then such independent reconstructions renders the sequence unrealistic, because, for example, a smiling face has very low probability of transitioning into a neutral face in next frame. Our intuition to tackle this problem is to constrain the possible models of generation by an auxiliary conditional information.  Such conditioning can be in the form of shape priors as used by Fi{\v{s}}er \textit{et al.} \cite{fivser2017example} for synthesis of stylized facial animations or consistency in optical flow field \cite{huang2017real} for video style transfer. IIzuka \textit{et al.} only concentrated on consistency of reconstruction at a local and global scale within a single frame \cite{iizuka2017globally}, but did not address the issue of multimodal image completion in sequence inpainting. We illustrate, both numerically and visually, the inconsistencies in GAN based inpainting methods and offer a simple yet computationally frugal solution to enforce consistency.
\par Regarding correctness: Correctness refers to a similarity metric quantifying the fidelity of reconstructed output to original version. As we are building upon the recent "DCGAN"\cite{dcgan} based inpainting method by Yeh \textit{et al.}\cite{yeh2017semantic}(we abbreviate this as `DIP' in rest of the paper),  the quality of reconstruction depends on the success of training the generator to approximate the underlying data distribution. Recent work by \cite{reed2016learning} shows that conditioning the GAN framework on positional constraints fosters in better sample generation. Our idea of improving upon \cite{yeh2017semantic} is to condition the GAN framework with automatically extracted facial semantics and thereby enabling (can be treated as constraining) the generator to generate specific facial components adhering to this conditioning input. In \textsection \ref{sec_turing}, we show that this simple yet elegant solution significantly improves quality of generated samples and also plays a pivotal role in achieving consistent reconstruction.  \\ 
Specifically, our key contributions in the paper are: 
\begin{enumerate}
\item To the best of our knowledge, this is the first time the dual concept of "correctness" and "consistency" is being explicitly studied in the context of GAN based inpainting. 
\item We present a novel semantically guided GAN architecture for generating more appealing images at 64$\times$64 and 128$\times$128 resolutions compared to \cite{yeh2017semantic,goodfellow2014generative}(\textsection \ref{sec_train_gan}). 
\item We show that the facial semantic conditional information enables our generative model to disentangle between appearance and pose cues (\textsection\ref{sec_independence}). 
\item A new framework is presented for assessing consistency of reconstruction by generative models. We show both that our model is able to reconstruct more consistent images compared to DIP (\textsection\ref{sec_consistency}).
\item We present extensive quantitative, qualitative and subjective evaluations to establish the superiority of samples generated by our proposed GAN model (\textsection\ref{sec_turing}) and subsequent inpainted reconstructions (\textsection \ref{sec_res_inpainting}).
\end{enumerate}
 \begin{figure*}[!t]
 \centering
 \includegraphics[scale = 0.26]{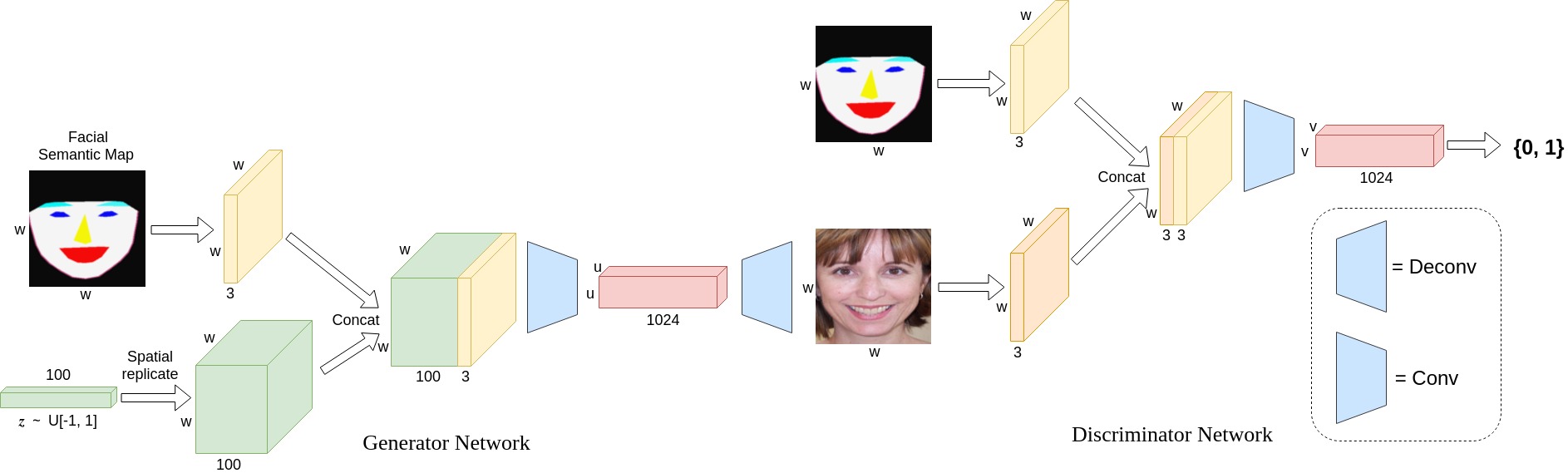}
 \caption{Proposed architecture of training semantically guided GAN. The generator network takes in a noise vector, $z$, an facial semantic map and generates a facial image. The discriminator is also conditioned on both real/generated images with corresponding facial maps and distinguishes between (image, map) pair belonging to real or generated distribution. The $z$ vector is spatially replicated to a spatial resolution of W$\times$W. Convolutional net of generator section reduces spatial resolution to $U=\frac{W}{16}$ . The  convolutional of Discriminator reduces spatial resolution of combined image and semantic map to, $U$ = 4$\times4$. Detailed explanation is in \textsection \ref{sec_train_gan} }
 \label{fig_architecture}
 \end{figure*}
\section{Related Works}
With the advent of Variational Auto-Encoder (VAE)\cite{kingma2013auto} and GAN \cite{goodfellow2014generative}, there has been a recent surge in interest towards automated image/video generation and subsequent unsupervised feature learning \cite{donahue2016adversarial,context_encoders,amader}. GANs are known to generate sharper images compared to VAE because VAE is based on the principle of $\ell_2$ loss based between generated images and posterior distribution and  thereby producing blurry outputs. In \cite{dcgan}, the authors recommend an empirically tested, stable architecture framework for GAN training and it became popular as the "DCGAN". While the original GAN formulation allowed the generator network to unrestrainedly sample from generated distribution, recently, researchers have utilized additional conditional information to constrain the output space of GANs for controlled sample generation. Class conditional GAN \cite{class_conditional} was the natural extension, wherein the generator was forced to generate samples of a given class. Denton \textit{et al.} \cite{denton2015deep} extended this idea in a class conditional Laplacian pyramid GAN setting. Such hierarchical conditioning information aided in better sample quality. Apart from discrete class labels, continuous attributes such as `smile', `age', etc., have been used in \cite{kaneko2017generative} to interactively modify a given image. Such continuous conditioning have also been leveraged by \cite{brock2016neural, zhu2016generative} for making semantically consistent photo editing on faces.
Conditioning on natural text was leveraged by Reed \textit{et al.} \cite{text2image} to directly map an informal description of a flower and bird to pixel space. Later, Zhang \textit{et al.} \cite{zhang2016stackgan} used a stacked GAN architecture to generate sharper images at higher resolution by conditioning on both text and first level of GAN generated image. 
\par Another contemporary practice is to condition a GAN on an auxiliary image, specially for the task of image-to-image translation
\cite{isola2016image, support1}, style transfer \cite{zhu2017unpaired,support2, johnson2016perceptual}, video/sequence generation \cite{mathieu2015deep, vondrick2016generating}, image denoising \cite{wolterink2017generative}, real time texture synthesis \cite{li2016precomputed}, image super resolution \cite{ledig2016photo}, semantic inpainting \cite{luc2016semantic}, unsupervised visual domain alignment \cite{bousmalis2016unsupervised,shrivastava2016learning} to list a few.
\par Conditional information is not only restricted to GANs. Recent works on VAE have also explored such auxiliary conditions for predicting future state from a single static image \cite{walker2016uncertain}, attribute based face editing \cite{yan2016attribute2image} and in learning to represent structured output \cite{sohn2015learning}. Conditional inputs have also been used as discriminative regularizers \cite{lamb2016discriminative} for improving VAE sample quality. 
\par Recently, Reed at al \cite{what_where} showed that providing sparse localization information to the generator network aids the generator in producing better samples. Our idea is mainly motivated from this observation. However, our approach is computationally more scalable because we use an automated facial fiducial points detection framework based on the real time face alignment with ensemble of regression trees \cite{kazemi2014one}. The authors in \cite{what_where} instead had to manually mark the  parts of the objects before training the conditional GAN. Also, we provide a dense semantic guide to the generator instead of sparse body joint or bounding box locations. 
\section{Preliminaries}
\subsection{Generative Adversarial Networks}
Generative adversarial network engages two parametrized models, viz., discriminator and generator in a two-player min-max game. Realized as a feed forward neural net, the generator network takes a latent noise vector $z$ drawn from a prior noise distribution $p_z(z)$. Following \cite{yeh2017semantic,goodfellow2014generative},  $z\sim \mathcal{U}[-1,1]$ (uniform distribution) and generator maps it onto an image, $y$; $G~: z~ \rightarrow y$. The other network, discriminator, has the task to discriminate samples coming from the true data distribution $p_D$ and the generated distribution, $p_G$. Specifically, generator and discriminator play the following game on $V(D, G)$:
$$\underset{G}{min}~~ \underset{D}{max}~~ V(D, G) = \mathbb{E}_{x\sim p_{data}(x)}[\log D(x)]$$
\begin{equation}
+ ~~\mathbb{E}_{z\sim p_{z}(z)}[1 - D(G(z))]
\end{equation}
This min-max game has global optimum when $p_{data}=p_G$ and this happens when both discriminator and generator have enough capacity \cite{goodfellow2014generative}. Empirically, it has been observed that for generator, it is prudent to maximize $\log (D(G(z)))$ instead of minimizing $\log [1 - D(G(z))] $.
\subsection{Conditional generative adversarial networks}
In conditional GANs, an extra input, $c$ is also fed to the generator in addition to the $z$ vector and thus $G~: (c \times z)~ \rightarrow y$. Under this conditioning, the modified objective for GAN training becomes,
$$\underset{G}{min}~~ \underset{D}{max}~~ V(D, G) = \mathbb{E}_{x,c\sim p_{data}(x,c)}[\log D(x,c)]$$
\begin{equation}
+ ~~\mathbb{E}_{z\sim p_{z}(z), c \sim p_{data}(c)}[1 - D(G(z,c))]
\end{equation}
The  GAN framework is flexible in accepting different genres of conditioning inputs such as class labels \cite{odena2016conditional}, natural language description \cite{text2image}, localization information\cite{what_where} and even an entire image \cite{zhang2016stackgan,isola2016image} or a sequence of images \cite{mathieu2015deep}. In our case, we condition the GAN framework with a facial semantic map capturing the pose(head orientation, size) and coarse facial expressions.
\section{Method}
\subsection{Facial Semantic Map Extraction} 
The first requirement to train our semantic guided GAN framework is to extract facial semantics. In that regard, we make use of the real time face alignment framework of Kazemi \textit{et al.} \cite{kazemi2014one}. However, detection of facial key points alone does not explicitly give semantic information of face. To mitigate this issue, semantically similar facial components are grouped together and given the same RGB color encoding.  As shown in Figure\ref{fig_architecture}, this semantic map acts an conditional information during GAN training and inference phases. 
\subsection{Training semantic conditioned GAN}\label{sec_train_gan}
The basic architecture of our proposed conditioned GAN training is shown in Figure \ref{fig_architecture}. We draw the noise prior $z \sim \mathcal{U}[-1,1]$ and tile to it all spatial locations to match the resolution of the conditional map. Next, the tiled $z$ vector and facial semantic maps are concatenated and fed to a conv-deconv \footnote{Deconv layer should ideally be termed as transposed convolution layer} network. The convolutional network has 5 layers of convolution of stride 2, kernel size 5 and number of filters doubles at every stage.  Next, the transposed convolutional section consists of 4 layers of fractionally strided convolution. Each layer upscales the previous layer's output by 2 and halves the number of filters.  The discriminator is also conditioned on the semantic map by concatenating the generated/real images with the corresponding maps. This forces the generator not only to generate realistic samples but also to adhere to the face pose and expressions constraints imposed by the semantic map. Discriminator consists of series of stride 2, kernel size 5 convolutions till the spatial resolution is reduced to 4$\times$4, followed by a linear layer which outputs the probability of joint  combination of (face, map) belonging to real/fake distribution. Following the recommendations in \cite{dcgan}, we apply Batch Normalization \cite{batchnorm} after all layers of the discriminator followed by ReLu non-linearity. Exception is the last deconvolutional layer which is followed by tanh non linearity without Batch Normalization.  In case of discriminator, except the first and last layer, Batch Normalization is applied after all the convolutional layers. We use LeakyReLu\cite{leakyrelu} non linearity activation after each convolutional layer. The final layer is followed by sigmoid non linearity.
\subsection{Semantic inpainting with appearance and pose constraint GAN} 
\label{sec_inpainting}
It was shown in \cite{dcgan} that a linear interpolation in the $z$ space results in smooth transition in semantic space. This indicates that semantically similar looking images can be created from `close' $z$ vectors (here, we define ``close'' as per Equation \ref{eq_total_loss}). We build upon the work of \cite{yeh2017semantic}, wherein the idea is to find the approximate $z$ vector related to the semantically ``closest'' natural image compared to the corrupted image. However, in our proposed case, the semantic closeness between corrupted and uncorrupted image is constrained by both appearance and pose criteria; such joint constraint helps in visually correct and structurally aligned inpainitng. Specifically, given a damaged image, $I^d$, the corruption mask, $D$, and the semantic map conditioning, $c$, we aim to find the best fit $z$ vector by iterative optimization of the following loss function,
\begin{equation}
    \hat{z} = \underset{z}{\mathrm{argmin}}~~ \{L_{con}(z|I^d,c,D) + \eta L_{per}(z|c)\}
    \label{eq_total_loss}
\end{equation}
where $\eta$ strikes a trade off between $L_{con}(z|I,c,D)$ and $L_{per}(z|c)$.
$L_{con}(\cdot)$ is the contextual loss which penalizes for changing the appearance of the uncorrupted pixels.
\begin{equation}
    L_{con}(z|I,c,D) = || D \odot G(z, c)  ||_1~, 
\end{equation}
where $\odot$ is the Hadamard product operator. $D(x, y) =1$ for uncorrupted pixels and 0 otherwise.
$L_{per}(\cdot)$ is the perceptual loss coming from the pre-trained discriminator of \textsection \ref{sec_train_gan} and penalizes if the joint combination of generated image and the semantic map lies away from the natural image manifold.
\begin{equation}
    L_{per}(z|c) = \log[1-(G(z, c))]
\end{equation}
For a given corrupted image, we start with a random $z \sim \mathcal{U}[-1,1]$ and iteratively update $z$ with stochastic gradient descent to minimize the loss in Equation \ref{eq_total_loss}. This enables us to approximately find the $\hat{z}$ vector which approximately maps the corrupted image to its closest semantic neighbor. After calculating $\hat{z}$, the inpainted image, $I^{inp}$, is formed by overlaying the corrupted image, $I^d$, with the reconstructed image, $G(\hat{z}|c)$. 
\begin{equation}
I^{inp} = D \odot I^d + (1-M) \ \odot G(\hat{z}, c)
\end{equation}
Authors in \cite{yeh2017semantic} reported that such overlaying leads to subtle local appearance mismatch and can be mitigated by post processing with Poisson blending \cite{perez2003poisson}.
\subsection{Implementation Details}
Both experiments of GAN training and inpainitng were performed on 64$\times$64 and 128$\times$128 resolutions.
We have followed  
\ mini-batch stochastic gradient descent optimization with mini-batch size of 64 using Adam \cite{adam} optimizer.   
During GAN training, learning rate was kept constant at 2$\times$10$^{-4}$ and Adam momentum parameters, $\beta_1$ and $\beta_2$ = 0.5. \\
During semantic inpainitng, learning rate was set to 5$\times$10$^{-2}$ and iterative back propagation was carried on for 1500 iterations, after which the loss in Equation \ref{eq_total_loss} saturates. $z$
vector was restricted to be within $[-1, 1]$ and $\eta$ was set to 0.1. Momentum parameters of Adam, were set to $\beta_1$=0.9 and $\beta_2$=0.99. Same parameters were used for both 64$\times64$ and 128$\times$128 resolutions and all deformation types. For the framework of DIP \cite{yeh2017semantic}, we have used the parameter settings as reported by the authors.
\subsection{Assumptions}
Our model assumes the presence of facial semantic map on a corrupted image. As of today, this is not an over restrictive assumption because current state-of-the-art facial landmark localizers \cite{sun2013deep} are able to perform appreciably under significant occlusions. Also, recent work of Luc \textit{et al.} \cite{semantic_predict} has shown significant promise of predicting future semantic maps of a scene. Thus, our assumption of having facial  is significantly pragmatic. Moreover, we envision that the concepts of this paper will be extended for video inpainting. In videos, temporal redundancy makes it possible to reuse facial maps from a preceding voxel. It is a frequent practice \cite{ebdelli2015video, kung2006spatial} in traditional video coding literature to reuse spatio-temporal information from nearby voxels for appearance and motion vector based error concealment. However, the main question we ask in this paper is, ``if \textit{somehow} we provide semantic maps to GAN, will that improve overall inpainting quality and consistency?''. Predicting facial semantic map under occlusion is an independent research and deviates readers from central theme of this paper.
\begin{figure}[!t]
\centering
\includegraphics[scale = 0.58]{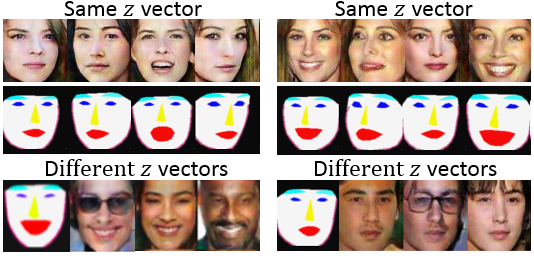}
\caption{Illustration of our proposed model learning to disentangle facial pose and appearance cues. \textbf{Top setting:} Faces generated with same $z$ vector but different semantic maps. \textbf{Bottom setting:} Faces generated with different $z$ vectors for a given semantic map. See supplementary document for more examples.}
\label{fig_pose_independence}
\end{figure}
\section{Experiments}
Before delving into experimental analysis, we feel, a justification is required for selecting DIP \cite{yeh2017semantic} as our comparing baseline and restricting ourselves to the original GAN formulation of Goodfellow \textit{et al.} \cite{goodfellow2014generative} for our GAN training. First, DIP, as of today, is the benchmark for GAN based image inpainitng. Superiority of GAN paradigm of inpainting over the previous state-of-the-art method of context encoders \cite{context_encoders} has already been shown in \cite{yeh2017semantic}. The core objective of this paper is to aware the readers of the intrinsic sequence inpainting inconsistency of DIP and to examine whether semantic conditioning aids in mitigating this drawback.
Second, we could have exploited other variants of GAN formulation such as Wasserstein GAN \cite{wgan}, boundary equilibrium GAN \cite{boundary} and unrolled GAN \cite{unrolled} as these variants have shown to produce better samples compared to original GAN formulation. However, in this paper we are interested in showing that conditional semantic map is effective to improve sample qualities compared to original unconstrained GAN formulation. Amalgamation of better GAN loss function and  semantic conditioning might not be a fair comparison to the framework of DIP.
 \begin{figure}[!t]
 \centering
 \includegraphics[scale = 0.6]{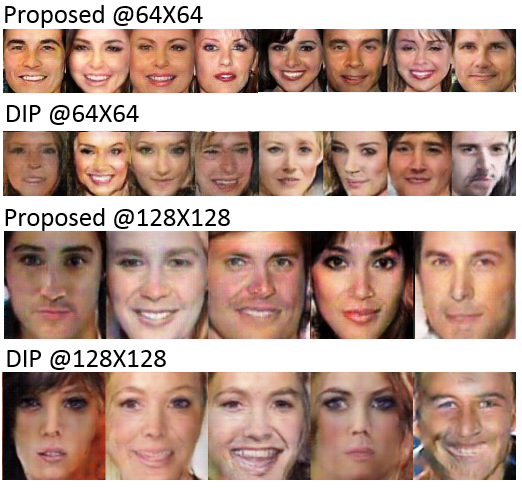}
 \caption{Visual comparison of random samples generated by our semantically conditioned GAN model and DIP\cite{yeh2017semantic}. Samples generated by our proposed network are superior in terms of visual quality (\textsection\ref{sec_turing}). See supplementary document for more examples.}
 \label{fig_gan_samples}
 \end{figure}
\subsection{Independence of appearance and pose}
\label{sec_independence}
Main hypothesis of our semantic conditioned GAN is that the generator should learn to disentangle appearance and pose cues for generating images. Intuitively, the semantic map should force the generator to create face with matching head pose and facial expression while two nearby $z$ vectors should result in similar facial textures. In Figure \ref{fig_pose_independence}(Top setting) we show groups of images which have been generated using same $z$ vectors but different semantic maps. Appearance factors such as gender, skin textures, hair color/styles are preserved; yet the facial expressions and pose closely adhere to the semantic map. In Figure \ref{fig_pose_independence}(bottom setting), images along a row are generated with different $z$ vectors for a given facial map. Changes in appearances can be appreciated but the facial expression/orientation remains constant. Such independence of appearance and shape is key in success of our inpainting method. Given a semantic map, the $z$ vector mainly focuses on perfecting the appearance. 
\subsection{Generated image quality and visual turing test}
\label{sec_turing}
Success of GAN based inpainting framework depends on the capability of the generator in approximating the real image manifold. So, a generator yielding more realistic samples is expected to perform better inpainting. Towards this end, we visually compare the  quality of random samples from our proposed semantic conditioned GAN and \cite{yeh2017semantic} at resolutions of 64$\times$64 and 128$\times$128. As shown in Figure \ref{fig_gan_samples}, samples from our proposed model are usually sharper and structurally more coherent. To quantitatively compare the visual appearance of the two models, we perform a visual turing test as followed in \cite{shrivastava2016learning}. A human annotator is randomly shown total 200 images(100 real and 100 generated) in groups of 20 and asked to label each sample as real or fake. Decisions from 10 annotators are taken. On average, at 64$\times64$ resolution, the classification accuracy is 5.8\% higher for DIP($p=10^{-3}$) and 4.2\% higher($p=10^{-2}$) at 128$\times$128 resolution. Thus, human annotators found it more difficult to distinguish samples from our dataset compared to DIP. This finding advocates the use of semantic conditioning for improving GAN samples without any significant overhead of architecture and loss function modification.
\begin{table}[!t]
\small
\centering
\caption{Mean correctness  of competing inpainting models measured in terms of PSNR (in dB) at 64$\times$64 and 128$\times$128 resolutions on CelebA test dataset. Ground truth images were corrupted with 4 types of masks, viz., Central, Checkboard, left and Freehand and fed to inpainting networks.}
\label{table_correctness} 
\begin{tabular}{lllll}
\hline
\multicolumn{5}{c}{\textbf{Resolution@64X64}}                  \\ \hline\hline
            & Central & Checkboard & Left  & Freehand   \\ \hline
DIP\cite{yeh2017semantic} & 24.96   & 18.51      & 17.13 & 24.53    \\ \hline
Proposed    & 26.32   & 19.97      & 18.42 & 25.86 \\ \hline
\multicolumn{5}{c}{\textbf{Resolution@128X128}}                  \\ \hline\hline
            & Central & Checkboard & Left  & Freehand     \\ \hline
DIP & 23.91   & 18.36     & 17.23 & 23.72    \\ \hline
Proposed    & 24.84   & 19.12      & 18.21 & 24.68 \\ \hline
\end{tabular}
\end{table}
 \subsection{Image inpainting }
\label{sec_res_inpainting}
As we interested in face inpainting, we have used the CelebA dataset which contains 202,599 face images. Following \cite{yeh2017semantic}, we separated 2000 images for testing. GAN training (both \cite{yeh2017semantic} and ours) was done on the remaining images.
 \begin{figure}[!h]
 \centering
 \includegraphics[scale = 0.55]{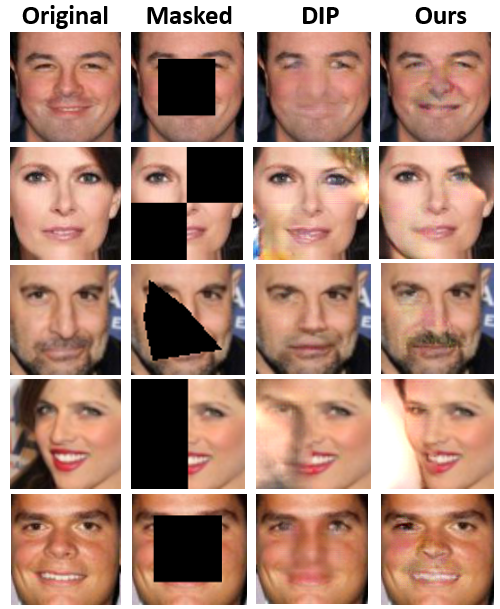}
 \caption{Inpainting comparison with DIP \cite{yeh2017semantic} at 128$\times$128 resolution. See supplementary document for more examples.}
 \label{fig_128}
 \end{figure}
\subsubsection{Correctness of inpainting: Quantitative and visual evaluation}
\begin{figure}[!t]
 \centering
 \includegraphics[scale = 0.65]{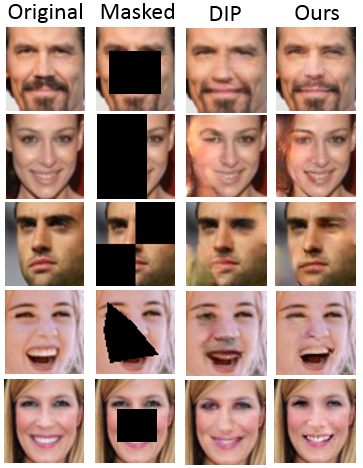}
 \caption{Inpainting comparison with DIP \cite{yeh2017semantic} at 64$\times$64 resolution. See supplementary document for more examples.}
 \label{fig_impaint_64}
 \end{figure}

For evaluating correctness, we measure the PSNR between an uncorrupted image and its inpainted version. While reporting PSNR, we have not used Poisson blending post processing because such post processing obscures the true performance of a generative deep model.
 We have performed extensive experiments on different types of corruption masks as shown in Figure \ref{fig_128} and \ref{fig_impaint_64} to compare the generalization capability of each model. The mean PSNR for each setting is reported in Table \ref{table_correctness}. At 64$\times$64 resolution for Central, Checkboard, Left and Freehand masks, our method outperforms DIP by margins of 1.36dB,  1.46dB, 1.29dB and 1.33 respectively. At 128$\times$128 resolution, corresponding margins are 0.93dB, 0.76dB, 0.98dB and 0.92dB.  Statistical significance of the observations reveals p-value $\leq 10^{-5}$ on all cases; this shows that our model significantly outperforms DIP. We show some exemplary inpaintings in Figure\ref{fig_128} and Figure\ref{fig_impaint_64}. It can be appreciated that finer facial structural details are preserved in our model. Our reconstructions are also sharper due to the intrinsic superiority of the underlying semantic conditioned GAN model.
 \par However, we acknowledge the fact that PSNR (or even Structural Similarity Metric \cite{wang2004image}(SSIM)) might not be the best metric to compare generative models because these models are not trained explicitly to minimize $\ell_2$ loss. Such observations were pointed out in recent works on image super resolution \cite{ledig2016photo, johnson2016perceptual}. To complement out findings in Table \ref{table_correctness}, we perform a human visual testing experiment. Each subject is shown the original image, the corrupted version and the two inpainted images without revealing the identity of the underlying algorithm. The subject has to vote for the inpainted image with better visual quality. Each subject was shown a random selection of 100 images. In a study with 10 participants, our algorithm selected 69.7\% of times which is significantly better ($p\leq 10^{-5}$) than chance.
 \subsection{Consistency in inpainting}
 \label{sec_consistency}
\begin{figure*}[!h]
 \centering
 \includegraphics[scale = 0.7]{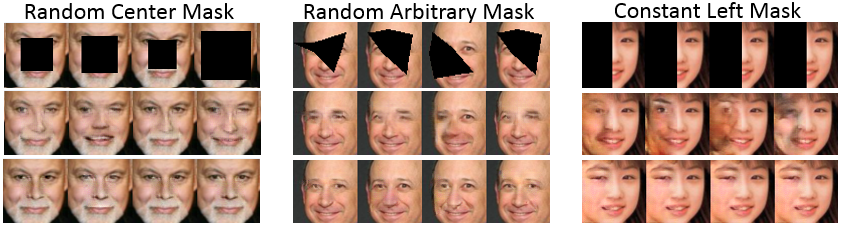}
 \caption{Comparison of consistent sequence inpainting. \textbf{Top row:} Pseudo sequence by corrupting a given image of CelebA dataset with random deformations. \textbf{Middle Row:} Inpainted output sequence by DIP \cite{yeh2017semantic}. \textbf{Bottom Row:} Inpainted output sequence by proposed model. Our model is more consistent in maintaining facial expressions and textures. See supplemental GIFs for better appreciation of consistency.}
 \label{fig_consistency}
 \end{figure*}

 The strategy behind enumerating consistency of inpainting is to corrupt a given image using different(or same) deformations and pass the corrupted images independently to the inpainting model. Ideally, each of the inpainted images should be coherent with each other. It is to be noted that such study of consistency is not recommended on real videos because there are unknown transformations between two successive frames. We can use an approximate motion compensation \cite{caballero2016real} to align two frames, but then the evaluation system will have an intrinsic motion compensation noise which is not separable from the generative modeling noise. Thus, we perform this study on pseudo sequences generated from the 2000 test images of CelebA. Examples of pseudo sequences are shown in Figure \ref{fig_consistency}.
 \par To formalize, given an uncorrupted image, $I^u$, we create a sequence comprising of $N$ different (or same) corrupted images, $I^u_{c_i}$ given by, $I^u_{c_i} = D_i(I_u)~ i \in \{1,2,.., N\}$; $D_i(\cdot)$ is corruption operator on $I^u$. Following Equation \ref{eq_total_loss}, for each $I^u_{c_i}$ we converge at a $\hat{z_i}$ and get the inpainted image, $G(\hat{z_i})$, from the generative model. For calculating consistency, we enumerate PSNR between all possible pairwise inpainted frames in the sequence. Consistency, $\eta^{Iu}$, for the pseudo sequence seeding from $I^u$, is calculated as,
\begin{table}[!t]
\small
\centering
\caption{Mean consistency (Refer to Equation\ref{eq_consistency}) on CelebA test set measured in terms of PSNR(in dB). A sequence was randomly perturbed by either one of Random Central, Random Freehand or constant 50\% Left masks. Higher consistency is better. }
\label{table_consistency}
\begin{tabular}{llll}
\hline
\multicolumn{4}{c}{\textbf{Resolution@64X64}}                                                                                                     \\ \hline\hline
         & \begin{tabular}[c]{@{}l@{}}Random\\ Central\end{tabular} & \begin{tabular}[c]{@{}l@{}}Random\\ Freehand\end{tabular} & Left  \\ \hline
DC-PAINT & 21.42 & 22.21 & 16.08 \\ \hline
Proposed & 26.14 & 26.15 & 16.72 \\ \hline
\multicolumn{4}{c}{\textbf{Resolution@128X128}}                                                                                                     \\ \hline\hline
         & \begin{tabular}[c]{@{}l@{}}Random\\ Central\end{tabular} & \begin{tabular}[c]{@{}l@{}}Random\\ Freehand\end{tabular} & Left  \\ \hline
DC-PAINT & 21.97 & 22.40 & 16.10 \\ \hline
Proposed & 23.81  & 25.60 & 17.15 \\ \hline
\end{tabular}
\end{table}
 \begin{equation}
     \eta^{I_u} =\frac{1}{\binom{N}{2}} \sum_{i=1}^N \sum_{j=1; j \neq i}^N~PSNR(G(\hat{z_i}),~G(\hat{z_j})) .
      \label{eq_consistency}
 \end{equation}
 \begin{figure*}[!h]
 \centering
 \includegraphics[scale = 0.47]{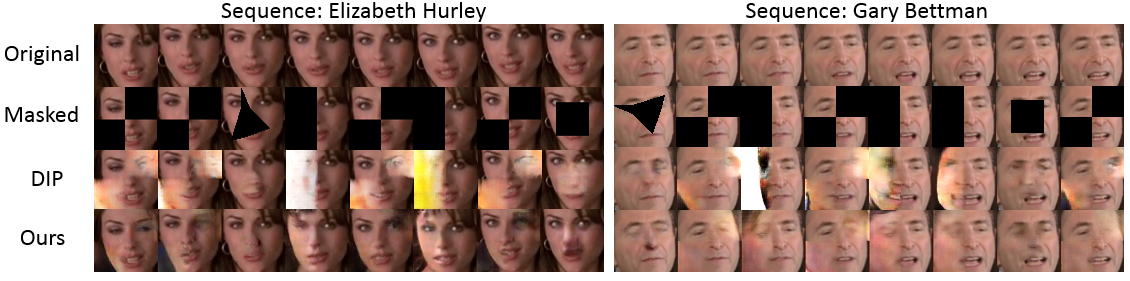}
 \caption{Visual comparison of inpainting on Youtube Faces video sequences.}
 \label{fig_youtube_face}
 \end{figure*}
 In Table \ref{table_consistency} we report the average values of consistency calculated over the 2000 sequences at 64$\times$64 and 128$\times$128 resolution with different deformation masks as shown in Figure \ref{fig_consistency}. Random center and Freehand masks depict the condition where each frame in sequence is corrupted by a different deformations (center mask varies from 50\%-70\%, Freehand mask corrupts random 25\% in 3 different freehand shapes). Constant left mask corrupts the left 50\% of a given frame for all frames in the pseudo sequence. 
  \par From Table \ref{table_consistency}, we see that our method is more consistent in reconstruction under all the different deformations. At 64$\times64$ resolution, for random central crop, mean consistency for our method is 4.62 dB higher than that of DIP . The corresponding margins are 3.94 dB  and 0.64 dB for random Freehand and left masks respectively.  At resolution of 128$\times$128, the average margins of success for our method are 1.84 dB, 3.2 dB and 1.05 dB respectively. Statistical significance of the observations reveals p-value $\leq 10^{-5}$ for each of the deformation settings and on both 64$\times$64 and 128$\times$128 resolution. Thus our proposed reconstructions are significantly more consistent than DIP.  ~To appreciate the numerical findings, we also show example cases in Figure \ref{fig_consistency}. It can be seen that DIP often fails to maintain consistency of the subtle yet important semantics such as facial expression (smile/neutral face), extent of eye opening, skin texture. Independently, each of the reconstructed frames by DIP might be acceptable, but when perceived as a sequence , the performance lacks realism due to abrupt change of facial semantics. Our model, however, faithfully retains not only the pose and expressions but also the skin texture of a subject. Such observation strongly bolsters our hypothesis that semantic guide is crucial in incorporating consistency in generative models.  
\begin{table}[tbp]
\centering
\caption{Comparison of PSNR (in dB) on Youtube Faces videos}
\label{table_youtube}
\begin{tabular}{lll}
\hline
Sequence         & DIP    & Ours  \\ \hline\hline
Gary Bettman     & 22.48 & 26.15 \\ \hline
Elizabeth Hurley & 13.05 & 23.67 \\ \hline
\end{tabular}
\end{table}
\section{Video inpainting on Youtube Faces}
As a proof of concept of viability of our model for video inpainting, we conducted preliminary experiments on the challenging Youtube Faces \cite{wolf2011face} dataset with the pretrained GAN models on CelebA. ~It is to be noted that in CelebA, the resolution of faces are bigger than in Youtube Faces dataset. The latter mainly captures celebrity videos in the wild. Thus there is an intrinsic domain difference between the distribution on which we trained GAN models and the distribution in which we are trying to do inpainting. As a result, visual quality of inpainting results are not at par with results on CelebA sequences. But, for this preliminary study, we were interested in examining the raw performances of the GAN models without any domain adaptation or retraining on Youtube videos. We randomly chose video sequences of 2 celebrities, viz. Elizabeth Hurley and Gary Bettman. We extracted the facial region from each frame, resized it to 64$\times$64 and corrupted randomly. The PSNR performance is reported in Table \ref{table_youtube} and a short snippet of qualitative visualizations are shown in Figure \ref{fig_youtube_face}. Even on real life video sequences, our model outperforms DIP, both visually and quantitatively. To our knowledge, this is the first time, a GAN based semantic inpainting framework has been applied on real life videos and our model shows a promising pathway in this regard.
\section{Discussion and Conclusion}
In this work we presented a simple yet effective framework for improving consistency and correctness in sequence inpainting by conditioning original GAN formulation with semantic mapping. Such conditioning also significantly improved the visual quality of generated samples both at 64$\times$64 and 128$\times$128. We showed that our model learns to disentangle appearance and pose information during sample generation and this helps in preserving both pose and appearance during inpainting. Also, we showed initial success of our model on Youtube Faces videos.
\par An important lesson here is that generative models are not naively suitable for video/sequence applications due to the multimodal nature of inference pipelines.  The results in this paper thus advocates the use of semantic inpainting for improving GAN performance, specially for video applications. In future we wish to study the combination of advanced variants of GAN \cite{wgan,boundary} with semantic conditioning. Another immediate extension would be to incorporate frameworks for predicting semantic mapping on corrupted portions of frames. 
{\small
\bibliographystyle{ieee}
\bibliography{egbib}

\begin{thebibliography}{10}\itemsep=-1pt

\bibitem{wgan}
M.~Arjovsky, S.~Chintala, and L.~Bottou.
\newblock Wasserstein generative adversarial networks.
\newblock In {\em ICML}, pages 214--223, 2017.

\bibitem{boundary}
D.~Berthelot, T.~Schumm, and L.~Metz.
\newblock Began: Boundary equilibrium generative adversarial networks.
\newblock {\em arXiv preprint arXiv:1703.10717}, 2017.

\bibitem{bousmalis2016unsupervised}
K.~Bousmalis, N.~Silberman, D.~Dohan, D.~Erhan, and D.~Krishnan.
\newblock Unsupervised pixel-level domain adaptation with generative
  adversarial networks.
\newblock {\em CVPR}, pages 3722--3731, 2017.

\bibitem{brock2016neural}
A.~Brock, T.~Lim, J.~M. Ritchie, and N.~Weston.
\newblock Neural photo editing with introspective adversarial networks.
\newblock {\em arXiv preprint arXiv:1609.07093}, 2016.

\bibitem{caballero2016real}
J.~Caballero, C.~Ledig, A.~Aitken, A.~Acosta, J.~Totz, Z.~Wang, and W.~Shi.
\newblock Real-time video super-resolution with spatio-temporal networks and
  motion compensation.
\newblock {\em CVPR}, 2016.

\bibitem{support2}
A.~J. Champandard.
\newblock Semantic style transfer and turning two-bit doodles into fine
  artworks.
\newblock {\em arXiv preprint arXiv:1603.01768}, 2016.

\bibitem{support1}
Q.~Chen and V.~Koltun.
\newblock Photographic image synthesis with cascaded refinement networks.
\newblock {\em arXiv preprint arXiv:1707.09405}, 2017.

\bibitem{denton2015deep}
E.~L. Denton, S.~Chintala, R.~Fergus, et~al.
\newblock Deep generative image models using a laplacian pyramid of adversarial
  networks.
\newblock In {\em NIPS}, pages 1486--1494, 2015.

\bibitem{donahue2016adversarial}
J.~Donahue, P.~Kr{\"a}henb{\"u}hl, and T.~Darrell.
\newblock Adversarial feature learning.
\newblock {\em arXiv preprint arXiv:1605.09782}, 2016.

\bibitem{ebdelli2015video}
M.~Ebdelli, O.~Le~Meur, and C.~Guillemot.
\newblock Video inpainting with short-term windows: application to object
  removal and error concealment.
\newblock {\em IEEE Transactions on Image Processing}, 24(10):3034--3047, 2015.

\bibitem{fivser2017example}
J.~Fi{\v{s}}er, O.~Jamri{\v{s}}ka, D.~Simons, E.~Shechtman, J.~Lu, P.~Asente,
  M.~Luk{\'a}{\v{c}}, and D.~S{\`y}kora.
\newblock Example-based synthesis of stylized facial animations.
\newblock {\em ACM Transactions on Graphics (TOG)}, 36(4):155, 2017.

\bibitem{goodfellow2014generative}
I.~Goodfellow, J.~Pouget-Abadie, M.~Mirza, B.~Xu, D.~Warde-Farley, S.~Ozair,
  A.~Courville, and Y.~Bengio.
\newblock Generative adversarial nets.
\newblock In {\em NIPS}, pages 2672--2680, 2014.

\bibitem{huang2017real}
H.~Huang, H.~Wang, W.~Luo, L.~Ma, W.~Jiang, X.~Zhu, Z.~Li, and W.~Liu.
\newblock Real-time neural style transfer for videos.
\newblock In {\em CVPR}, pages 783--791, 2017.

\bibitem{iizuka2017globally}
S.~Iizuka, E.~Simo-Serra, and H.~Ishikawa.
\newblock Globally and locally consistent image completion.
\newblock {\em ACM Transactions on Graphics (TOG)}, 36(4):107, 2017.

\bibitem{batchnorm}
S.~Ioffe and C.~Szegedy.
\newblock Batch normalization: Accelerating deep network training by reducing
  internal covariate shift.
\newblock In {\em ICML}, pages 448--456, 2015.

\bibitem{isola2016image}
P.~Isola, J.-Y. Zhu, T.~Zhou, and A.~A. Efros.
\newblock Image-to-image translation with conditional adversarial networks.
\newblock {\em CVPR}, pages 1125--1134, 2016.

\bibitem{johnson2016perceptual}
J.~Johnson, A.~Alahi, and L.~Fei-Fei.
\newblock Perceptual losses for real-time style transfer and super-resolution.
\newblock In {\em ECCV}, pages 694--711. Springer, 2016.

\bibitem{kaneko2017generative}
T.~Kaneko, K.~Hiramatsu, and K.~Kashino.
\newblock Generative attribute controller with conditional filtered generative
  adversarial networks.
\newblock In {\em CVPR}, pages 6089--6098, 2017.

\bibitem{kazemi2014one}
V.~Kazemi and J.~Sullivan.
\newblock One millisecond face alignment with an ensemble of regression trees.
\newblock In {\em CVPR}, pages 1867--1874, 2014.

\bibitem{adam}
D.~Kingma and J.~Ba.
\newblock Adam: A method for stochastic optimization.
\newblock {\em arXiv preprint arXiv:1412.6980}, 2014.

\bibitem{kingma2013auto}
D.~P. Kingma and M.~Welling.
\newblock Auto-encoding variational bayes.
\newblock {\em arXiv preprint arXiv:1312.6114}, 2013.

\bibitem{kung2006spatial}
W.-Y. Kung, C.-S. Kim, and C.-C. Kuo.
\newblock Spatial and temporal error concealment techniques for video
  transmission over noisy channels.
\newblock {\em IEEE transactions on circuits and systems for video technology},
  16(7):789--803, 2006.

\bibitem{amader}
A.~Lahiri, K.~Ayush, P.~K. Biswas, and P.~Mitra.
\newblock Generative adversarial learning for reducing manual annotation in
  semantic segmentation on large scale miscroscopy images: Automated vessel
  segmentation in retinal fundus image as test case.
\newblock In {\em Proceedings of the IEEE Conference on Computer Vision and
  Pattern Recognition Workshops}, pages 42--48, 2017.

\bibitem{lamb2016discriminative}
A.~Lamb, V.~Dumoulin, and A.~Courville.
\newblock Discriminative regularization for generative models.
\newblock {\em arXiv preprint arXiv:1602.03220}, 2016.

\bibitem{ledig2016photo}
C.~Ledig, L.~Theis, F.~Husz{\'a}r, J.~Caballero, A.~Cunningham, A.~Acosta,
  A.~Aitken, A.~Tejani, J.~Totz, Z.~Wang, et~al.
\newblock Photo-realistic single image super-resolution using a generative
  adversarial network.
\newblock {\em CVPR}, pages 4681--4690, 2016.

\bibitem{li2016precomputed}
C.~Li and M.~Wand.
\newblock Precomputed real-time texture synthesis with markovian generative
  adversarial networks.
\newblock In {\em ECCV}, pages 702--716. Springer, 2016.

\bibitem{luc2016semantic}
P.~Luc, C.~Couprie, S.~Chintala, and J.~Verbeek.
\newblock Semantic segmentation using adversarial networks.
\newblock {\em NIPS Workshop on Adversarial Learning}, 2016.

\bibitem{leakyrelu}
A.~L. Maas, A.~Y. Hannun, and A.~Y. Ng.
\newblock Rectifier nonlinearities improve neural network acoustic models.
\newblock In {\em ICML}, volume~30, 2013.

\bibitem{mathieu2015deep}
M.~Mathieu, C.~Couprie, and Y.~LeCun.
\newblock Deep multi-scale video prediction beyond mean square error.
\newblock {\em ICLR}, 2016.

\bibitem{unrolled}
L.~Metz, B.~Poole, D.~Pfau, and J.~Sohl-Dickstein.
\newblock Unrolled generative adversarial networks.
\newblock {\em arXiv preprint arXiv:1611.02163}, 2016.

\bibitem{class_conditional}
M.~Mirza and S.~Osindero.
\newblock Conditional generative adversarial nets.
\newblock {\em arXiv preprint arXiv:1411.1784}, 2014.

\bibitem{semantic_predict}
N.~Neverova, P.~Luc, C.~Couprie, J.~Verbeek, and Y.~LeCun.
\newblock Predicting deeper into the future of semantic segmentation.
\newblock {\em arXiv preprint arXiv:1703.07684}, 2017.

\bibitem{odena2016conditional}
A.~Odena, C.~Olah, and J.~Shlens.
\newblock Conditional image synthesis with auxiliary classifier {GAN}s.
\newblock In {\em ICML}, pages 2642--2651, 2017.

\bibitem{context_encoders}
D.~Pathak, P.~Krahenbuhl, J.~Donahue, T.~Darrell, and A.~A. Efros.
\newblock Context encoders: Feature learning by inpainting.
\newblock In {\em CVPR}, pages 2536--2544, 2016.

\bibitem{perez2003poisson}
P.~P{\'e}rez, M.~Gangnet, and A.~Blake.
\newblock Poisson image editing.
\newblock In {\em ACM Transactions on graphics (TOG)}, volume~22, pages
  313--318. ACM, 2003.

\bibitem{dcgan}
A.~Radford, L.~Metz, and S.~Chintala.
\newblock Unsupervised representation learning with deep convolutional
  generative adversarial networks.
\newblock {\em ICLR}, 2016.

\bibitem{text2image}
S.~Reed, Z.~Akata, X.~Yan, L.~Logeswaran, B.~Schiele, and H.~Lee.
\newblock Generative adversarial text-to-image synthesis.
\newblock In {\em ICML}, 2016.

\bibitem{reed2016learning}
S.~E. Reed, Z.~Akata, S.~Mohan, S.~Tenka, B.~Schiele, and H.~Lee.
\newblock Learning what and where to draw.
\newblock In {\em NIPS}, pages 217--225, 2016.

\bibitem{what_where}
S.~E. Reed, Z.~Akata, S.~Mohan, S.~Tenka, B.~Schiele, and H.~Lee.
\newblock Learning what and where to draw.
\newblock In {\em NIPS}, pages 217--225, 2016.

\bibitem{shrivastava2016learning}
A.~Shrivastava, T.~Pfister, O.~Tuzel, J.~Susskind, W.~Wang, and R.~Webb.
\newblock Learning from simulated and unsupervised images through adversarial
  training.
\newblock {\em CVPR}, pages 2107--2116, 2017.

\bibitem{sohn2015learning}
K.~Sohn, H.~Lee, and X.~Yan.
\newblock Learning structured output representation using deep conditional
  generative models.
\newblock In {\em NIPS}, pages 3483--3491, 2015.

\bibitem{sun2013deep}
Y.~Sun, X.~Wang, and X.~Tang.
\newblock Deep convolutional network cascade for facial point detection.
\newblock In {\em CVPR}, pages 3476--3483, 2013.

\bibitem{vondrick2016generating}
C.~Vondrick, H.~Pirsiavash, and A.~Torralba.
\newblock Generating videos with scene dynamics.
\newblock In {\em NIPS}, pages 613--621, 2016.

\bibitem{walker2016uncertain}
J.~Walker, C.~Doersch, A.~Gupta, and M.~Hebert.
\newblock An uncertain future: Forecasting from static images using variational
  autoencoders.
\newblock In {\em ECCV}, pages 835--851. Springer, 2016.

\bibitem{wang2004image}
Z.~Wang, A.~C. Bovik, H.~R. Sheikh, and E.~P. Simoncelli.
\newblock Image quality assessment: from error visibility to structural
  similarity.
\newblock {\em IEEE transactions on image processing}, 13(4):600--612, 2004.

\bibitem{wolf2011face}
L.~Wolf, T.~Hassner, and I.~Maoz.
\newblock Face recognition in unconstrained videos with matched background
  similarity.
\newblock {\em CVPR}, pages 529--534, 2011.

\bibitem{wolterink2017generative}
J.~M. Wolterink, T.~Leiner, M.~A. Viergever, and I.~Isgum.
\newblock Generative adversarial networks for noise reduction in low-dose ct.
\newblock {\em IEEE Transactions on Medical Imaging}, 2017.

\bibitem{yan2016attribute2image}
X.~Yan, J.~Yang, K.~Sohn, and H.~Lee.
\newblock Attribute2image: Conditional image generation from visual attributes.
\newblock In {\em ECCV}, pages 776--791. Springer, 2016.

\bibitem{yeh2017semantic}
R.~A. Yeh, C.~Chen, T.~Y. Lim, A.~G. Schwing, M.~Hasegawa-Johnson, and M.~N.
  Do.
\newblock Semantic image inpainting with deep generative models.
\newblock In {\em CVPR}, pages 5485--5493, 2017.

\bibitem{zhang2016stackgan}
H.~Zhang, T.~Xu, H.~Li, S.~Zhang, X.~Huang, X.~Wang, and D.~Metaxas.
\newblock Stackgan: Text to photo-realistic image synthesis with stacked
  generative adversarial networks.
\newblock {\em CVPR}, pages 5077--5086, 2016.

\bibitem{zhu2016generative}
J.-Y. Zhu, P.~Kr{\"a}henb{\"u}hl, E.~Shechtman, and A.~A. Efros.
\newblock Generative visual manipulation on the natural image manifold.
\newblock In {\em ECCV}, pages 597--613. Springer, 2016.

\bibitem{zhu2017unpaired}
J.-Y. Zhu, T.~Park, P.~Isola, and A.~A. Efros.
\newblock Unpaired image-to-image translation using cycle-consistent
  adversarial networks.
\newblock {\em arXiv preprint arXiv:1703.10593}, 2017.

\end{thebibliography}
}
\cleardoublepage
\newpage

\title{Additional Results}
\date{}
\author{}
\maketitle

\section{Visualization: Independence of appearance and pose}
\begin{figure*}[!t]
 \centering
 \includegraphics[scale=0.99]{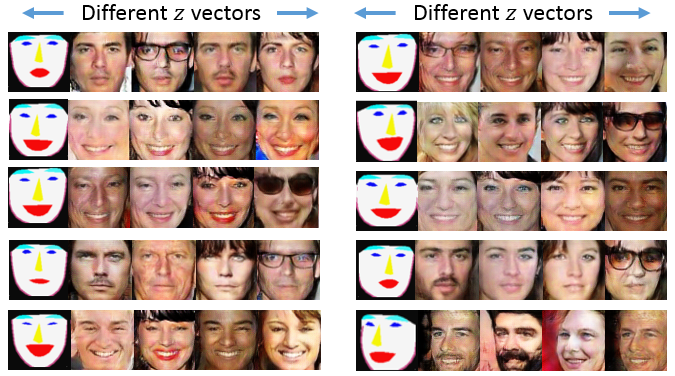}
 \caption{Faces generated for different $z$ vectors but a given semantic facial map. Each set of images is generated using different $z$ vectors but conditioned on a given facial map. Note, how the appearance of the faces changes across the columns for a given set, but the facial expression and orientation is modulated by the conditioning map.}
 \label{fig_same_key_diff_z}
 \end{figure*}

\begin{figure*}[!t]
 \centering
 \includegraphics[scale = .99]{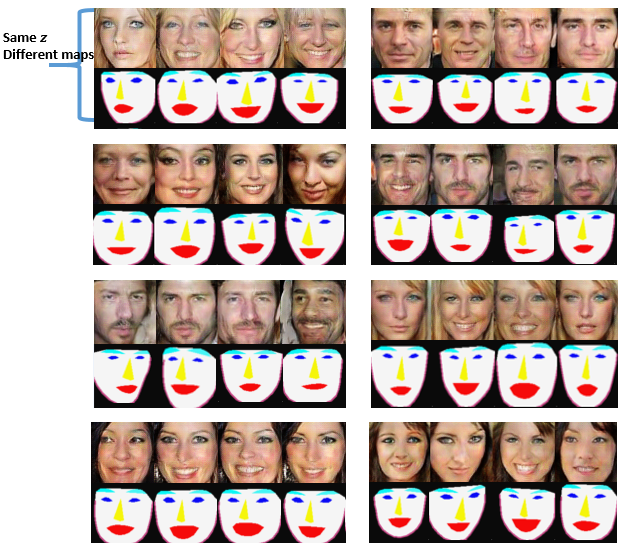}
 \caption{Faces generated for different facial semantic maps but a given $z$ vector. Each set of faces is created with same $z$ vector but different facial semantic map. Note, how facial expressions and orientations are modulated by conditioning maps but facial appearance cues such as texture, gender, hair color etc., are preserved.}
 \label{fig_same_z_diff_k}
 \end{figure*}

As mentioned in Section 5.1 of main paper, independence of appearance and pose refers to the fact that our generator learns to disentangle appearance and pose cues for generating images. In Figure \ref{fig_same_key_diff_z}, we show example cases in which different faces are generated for a given semantic map. In Figure \ref{fig_same_z_diff_k}, we show example cases in which similar looking faces are generated from the same $z$ vector but conditioned on different facial maps.
 
\section{Visualization : Inpainting performance}
\begin{figure*}[!t]
 \centering
 \includegraphics[scale = .99]{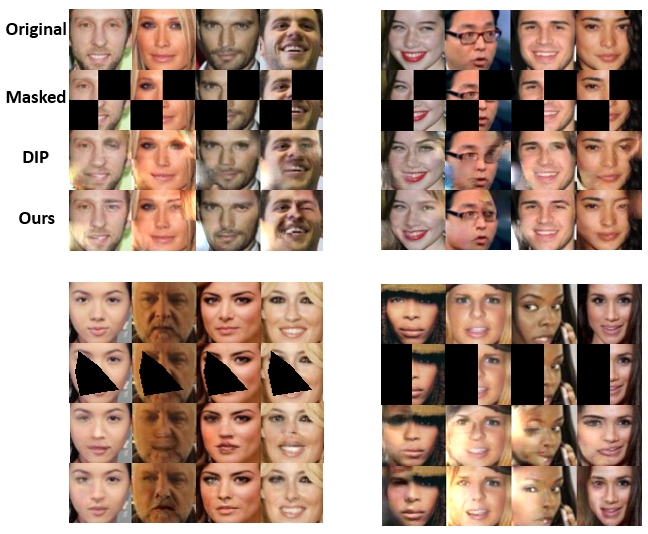}
 \caption{Comparison of inpainting at 64$\times$64 resolution with DIP \cite{yeh2017semantic}. \textbf{1$^{st}$ row:} Original image; \textbf{2$^{nd}$} row: Damaged image; \textbf{3$^{rd}$ row:} Inpainting by DIP; \textbf{4$^{th}$ row:} Inpainting by our method.}
 \label{fig-complete-64-1}
 \end{figure*}

\begin{figure*}[!t]
 \centering
 \includegraphics[scale = .99]{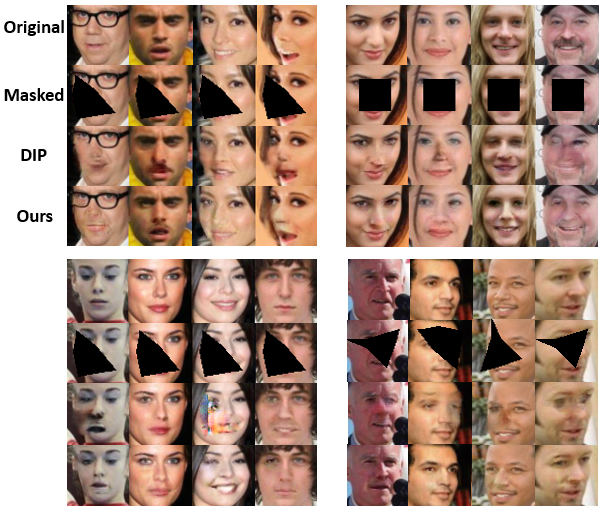}
 \caption{Comparison of inpainting at 64$\times$64 resolution with DIP \cite{yeh2017semantic}. \textbf{1$^{st}$ row:} Original image; \textbf{2$^{nd}$} row: Damaged image; \textbf{3$^{rd}$ row:} Inpainting by DIP; \textbf{4$^{th}$ row:} Inpainting by our method.}
 \label{fig-complete-64-2}
 \end{figure*}

\begin{figure*}[!t]
 \centering
 \includegraphics[scale = .65]{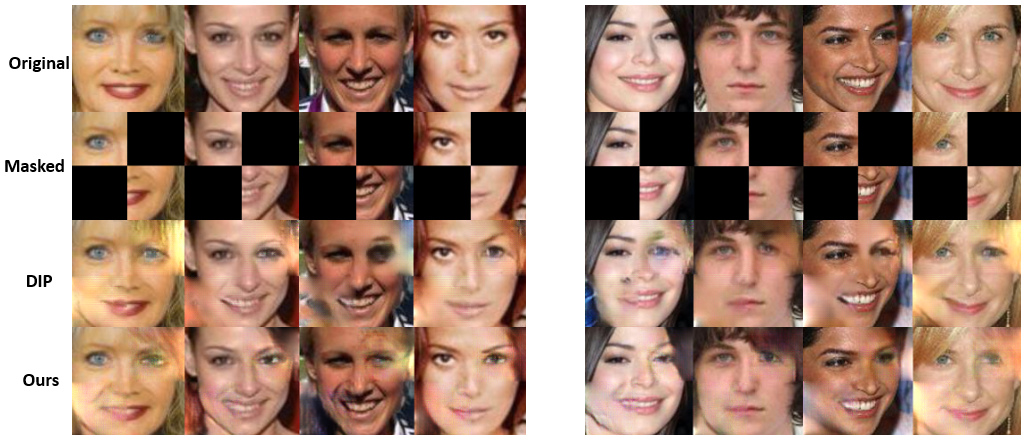}
 \caption{Comparison of inpainting at 128$\times$128 resolution with DIP \cite{yeh2017semantic}. \textbf{1$^{st}$ row:} Original image; \textbf{2$^{nd}$} row: Damaged image; \textbf{3$^{rd}$ row:} Inpainting by DIP; \textbf{4$^{th}$ row:} Inpainting by our method.}
 \label{fig-complete-128-1}
 \end{figure*}

\begin{figure*}[!t]
 \centering
 \includegraphics[scale = .65]{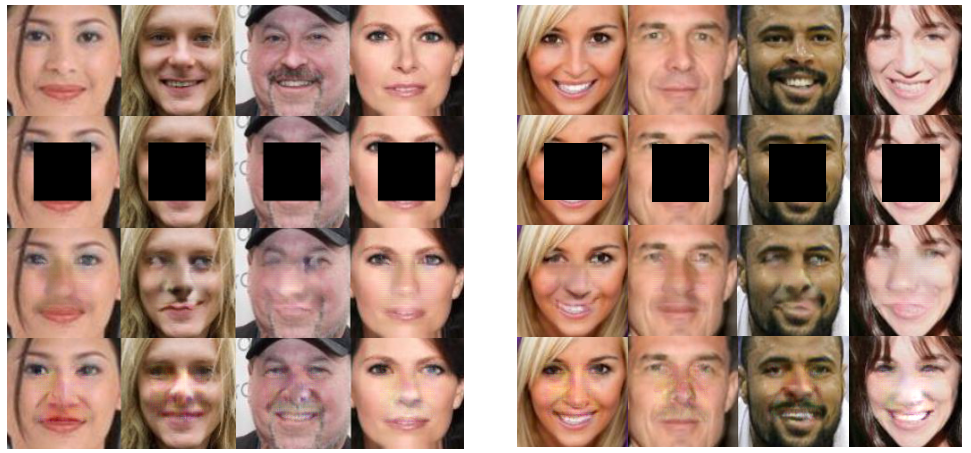}
 \caption{Comparison of inpainting at 128$\times$128 resolution with DIP \cite{yeh2017semantic}. \textbf{1$^{st}$ row:} Original image; \textbf{2$^{nd}$} row: Damaged image; \textbf{3$^{rd}$ row:} Inpainting by DIP; \textbf{4$^{th}$ row:} Inpainting by our method.}
 \label{fig-complete-128-2}
 \end{figure*}

In Figures \ref{fig-complete-64-1} and \ref{fig-complete-64-2} we visually compare some cases of semantic impainting on CelebA dataset at 64$\times$64 resolution by our model and DIP \cite{yeh2017semantic}. In Figures \ref{fig-complete-128-1} and \ref{fig-complete-128-2} we show comparisons of inpainting at 128$\times$128 resolution.

\section{Visualization : Consistency of inpainting}
\begin{figure*}[!t]
 \centering
 \includegraphics[scale = .99]{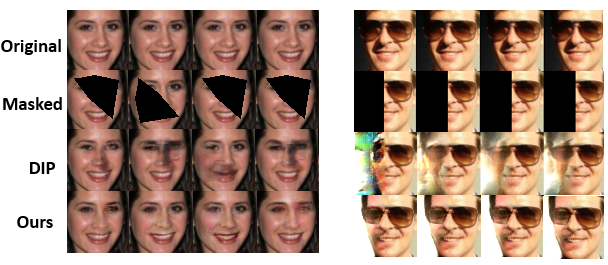}
 \caption{Comparison of consistency of inpainting with DIP \cite{yeh2017semantic} at 64$\times$64 resolution. \textbf{1$^{st}$ row:} Original image; \textbf{2$^{nd}$} row: Damaged image; \textbf{3$^{rd}$ row:} Inpainting by DIP; \textbf{4$^{th}$ row:} Inpainting by our method. It can be appreciated visually that our method is more consistent. }
 \label{fig_consistency-64}
 \end{figure*}

\begin{figure*}[!b]
 \centering
 \includegraphics[scale = .99]{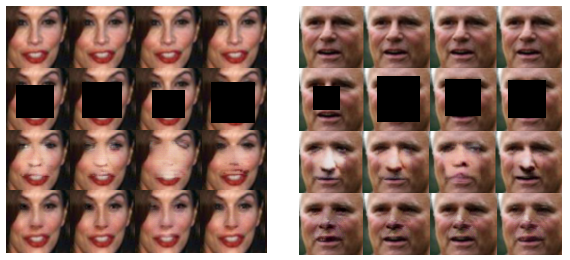}
 \caption{Comparison of consistency of inpainting with DIP \cite{yeh2017semantic} at 64$\times$64 resolution. \textbf{1$^{st}$ row:} Original image; \textbf{2$^{nd}$} row: Damaged image; \textbf{3$^{rd}$ row:} Inpainting by DIP; \textbf{4$^{th}$ row:} Inpainting by our method. It can be appreciated visually that our method is more consistent. }
 \label{fig_consistency-64-2}
 \end{figure*}

\begin{figure*}[!b]
 \centering
 \includegraphics[scale = .65]{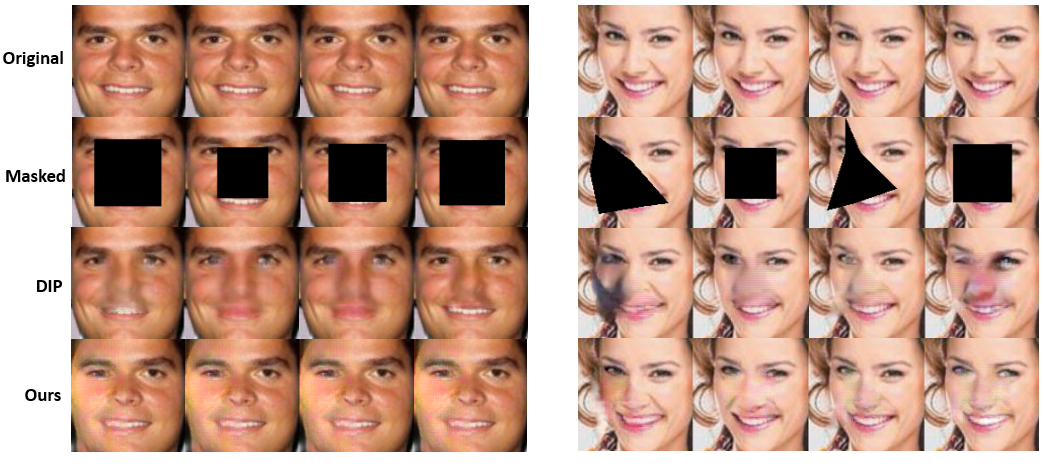}
 \caption{Comparison of consistency of inpainting with DIP \cite{yeh2017semantic} at 128$\times$128 resolution. \textbf{1$^{st}$ row:} Original image; \textbf{2$^{nd}$} row: Damaged image; \textbf{3$^{rd}$ row:} Inpainting by DIP; \textbf{4$^{th}$ row:} Inpainting by our method. It can be appreciated visually that our method is more consistent. }
 \label{fig_consistency-128-1}
 \end{figure*}

\begin{figure*}[!b]
 \centering
 \includegraphics[scale = .65]{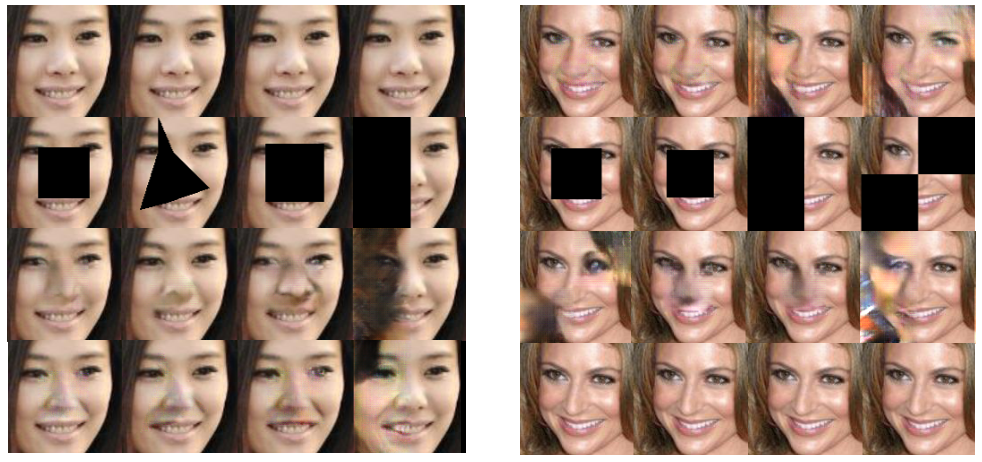}
 \caption{Comparison of consistency of inpainting with DIP \cite{yeh2017semantic} at 128$\times$128 resolution. \textbf{1$^{st}$ row:} Original image; \textbf{2$^{nd}$} row: Damaged image; \textbf{3$^{rd}$ row:} Inpainting by DIP; \textbf{4$^{th}$ row:} Inpainting by our method. It can be appreciated visually that our method is more consistent. }
 \label{fig_consistency-128-2}
 \end{figure*}

In Figures \ref{fig_consistency-64} and \ref{fig_consistency-64-2} we visually compare consistency of inpainting of a pseudo sequences at 64$\times$64 resolution.
Figures \ref{fig_consistency-128-1} and \ref{fig_consistency-128-2} show the visualizations for 128$\times$128 resolution.
Given an uncorrupted image, a pseudo sequence is created by deforming the original image with different(or same) masks. Refer to Section 5.4 in main paper for more details. 

\section{Visualization: Quality of generated samples from GAN }
\begin{figure*}[]
 \centering
 \includegraphics[scale = .80]{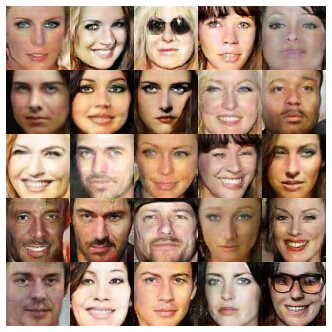}
 \caption{Samples from our proposed semantic conditioned GAN model at 64$\times$64 resolution. Visual quality of samples is better than those produced by DIP \cite{yeh2017semantic}.}
 \label{fig_proposed_gan_64}
 \end{figure*}

\begin{figure*}[]
 \centering
 \includegraphics[scale = .8]{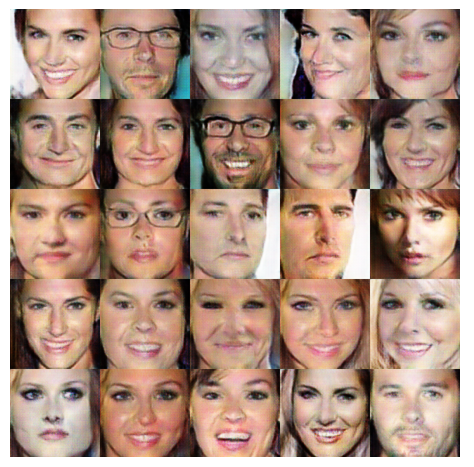}
 \caption{Samples from our proposed semantic conditioned GAN model at 128$\times$128 resolution. Visual quality of samples is better than those produced by DIP \cite{yeh2017semantic}.}
 \label{fig_proposed_gan_128}
 \end{figure*}

\begin{figure*}[]
 \centering
 \includegraphics[scale = .80]{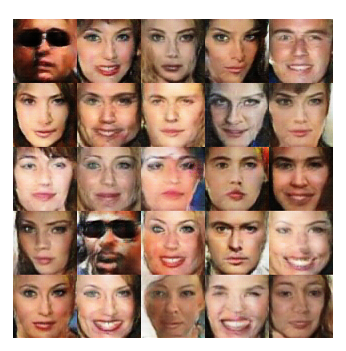}
 \caption{Samples from DCGAN model followed by \cite{yeh2017semantic} at 64$\times$64 resolution.}
 \label{fig_dcgan_64}
 \end{figure*}

\begin{figure*}[]
 \centering
 \includegraphics[scale = .8]{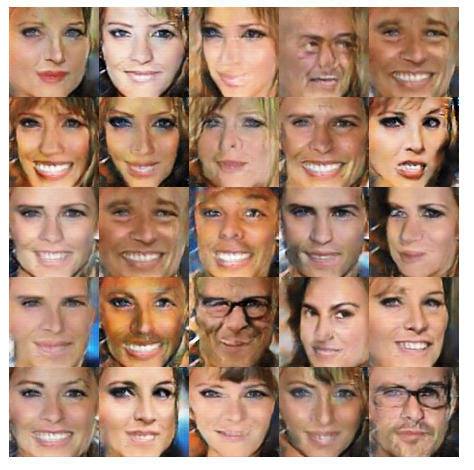}
 \caption{Samples from DCGAN model followed by \cite{yeh2017semantic} at 128$\times$128 resolution.}
 \label{fig_dcgan_128}
 \end{figure*}

In Figures \ref{fig_proposed_gan_64} and \ref{fig_proposed_gan_128} we show some samples generated by our semantically conditioned GAN at 64$\times$64 and 128$\times$128 resolution respectively. In Figures \ref{fig_dcgan_64} and \ref{fig_dcgan_128} the corresponding samples from DCGAN architecture used by DIP are shown. Qualitatively, sample qualities from our GAN model is better. Refer to Section 5.2 of main paper for a detailed analysis.

\end{document}